\documentclass[runningheads]{llncs}

\usepackage[utf8]{inputenc}
\usepackage{graphicx}
\usepackage{url, color}
\usepackage{hyperref}
\usepackage[nameinlink]{cleveref}
\usepackage{csquotes}
\usepackage{listings}
\usepackage{xspace}
\usepackage{lscape}

\usepackage{array}
\newcolumntype{L}[1]{>{\raggedright\let\newline\\\arraybackslash\hspace{0pt}}m{#1}}
\newcolumntype{C}[1]{>{\centering\let\newline\\\arraybackslash\hspace{0pt}}m{#1}}
\newcolumntype{R}[1]{>{\raggedleft\let\newline\\\arraybackslash\hspace{0pt}}m{#1}}

\newcommand{\essence}[0]{\textsc{Essence}}
\newcommand{\conjure}{\textsc{Conjure}\xspace}
\newcommand{\savilerow}{\textsc{Savile Row}\xspace}

\newcommand{\minion}{\textsc{Minion}\xspace}

\begin{document}

\title{Modelling Langford's Problem: \texorpdfstring{\newline}{} A Viewpoint for Search}
\author{Özgür Akgün \and Ian Miguel}
\institute{School of Computer Science, University of St Andrews \email{\{ozgur.akgun,ijm\}@st-andrews.ac.uk}}


\maketitle

\begin{abstract}

The performance of enumerating all solutions to an instance of Langford's Problem is sensitive to the model and the search strategy. In this paper we compare the performance of a large variety of models, all derived from two base viewpoints. We empirically show that a channelled model with a static branching order on one of the viewpoints offers the best performance out of all the options we consider. Surprisingly, one of the base models proves very effective for propagation, while the other provides an effective means of stating a static search order.

% \keywords{First keyword  \and Second keyword \and Another keyword.}
\end{abstract}

\section{Introduction}

Langford's Problem (number 24 at www.csplib.org) is specified as follows.

\begin{displayquote}

Arrange $k$ sets of numbers 1 to $n$ in a sequence, so that each appearance of the number $m$ is $m$ numbers on from the last. For example, the L(3,9) problem is to arrange 3 sets of the numbers 1 to 9 so that the first two 1's and the second two 1's appear one number apart, the first two 2's and the second two 2's appear two numbers apart, etc.

\end{displayquote}

Alternative constraint programming models and search strategies for this problem were explored first by Smith \cite{smith2000modelling}. Smith presented two models: the first corresponds to what we call the \emph{Positional} approach in this paper (\Cref{section-positional-model}), and the second model is the dual of the first one. When viewed as a permutation problem, the dual viewpoint for a permutation may be constructed by interchanging the variables and the values. This approach has been studied further and generalised by \cite{walsh2001permutation,smith2001dual,hnich2003channel,hnich2004dual,smith9comparing}. Smith's second model corresponds to the \emph{Direct} approach in this paper (\Cref{section-direct-model}) with one exception. In our direct approach, a value $m$ is repeated $k$ times, whereas in \cite{smith2000modelling} the second occurrence of a value $m$ is represented with $m+n$, the third with $m+2n$, and so on.

In this paper we investigate the performance of a number of different models and search strategies for Langford's Problem. We first present two approaches to modelling Langford's Problem, using two separate viewpoints. We then employ a very strict channelling constraint between the two viewpoints to construct a channelled approach. We compare 20 variations in total, including the best search heuristic identified by \cite{smith2000modelling} and the AllSAT solver \texttt{BC\_MINISAT\_ALL} \cite{toda2016implementing} to enumerate all solutions. Our empirical evaluation shows that a static branching order on the Direct viewpoint in a channelled model offers the best performance.

% {\bf Note}. Here, some more citations: \cite{kiziltan2001symmetry,choi2003propagation,choi2003propagation2,choi2007removing,law2007automatic,smith9comparing}.

\section{Two Approaches to Modelling Langford's Problem}

In what follows we present two approaches to modelling this problem, using two separate viewpoints, in the \essence{} \cite{frisch2005essence,frisch2007design,frisch2008ssence} constraint specification language. One is direct, has a sequence of numbers just like the problem specification. The other is positional, has a function from numbers and the repetition to the position in the sequence. 

\subsection{A \emph{Direct} Approach \label{section-direct-model}}

We first present a \emph{Direct} problem specification for Langford's Problem, given in \Cref{direct-model}. An \essence{} specification identifies: the input parameters of the problem class ({\tt given}), whose values define an instance; the combinatorial objects to be found ({\tt find}); the constraints the objects must satisfy ({\tt such that}); identifiers declared ({\tt letting}); and an (optional) objective function ({\tt min/maximising}). This specification contains a single decision variable with a domain of type \texttt{sequence}, with a fixed length. In order to solve an \essence{} specification, it is first refined into a solver-independent constraint model using the \conjure{} automated constraint modelling system
\cite{akgun2014extensible,akgun2013automated,akgun2014breaking,akgun2011extensible} and then prepared for input to a particular constraint solver, such as {\sc Minion} \cite{gent2006minion}, or SAT by the \savilerow{} system \cite{nightingale2014automatically,sr-journal-17,nightingale2015automatically}.

Sequences in \essence{} are not required to be of fixed length in general. A fixed length sequence is very similar to a one-dimensional array, except for the provision of some additional operators. For example, in this specification we use the \texttt{preImage} operator (on line 29) to denote all the index positions of a certain number. The apartness constraint (on lines 11--21) is written as a triply-nested quantified expression. For each number there exists a starting position: the first index where we see this particular number in the sequence. Once this position is denoted, the apartness constraint can be posted by constraining the appropriate positions in the sequence to be equal to the number in focus. As an alternative to the existential quantification we could have used auxiliary variables to denote the first positions of each number as a separate \essence{} variable. However, in this case we feel the existential quantification is clearer. Furthermore, our constraint modelling pipeline of \conjure{} and \savilerow{} is able to generate a top level decision variable for the existentially quantified variables automatically. The second constraint (on line 24) is for breaking the reflection symmetry. The third constraint (lines 26--29) is implied: it states the fact that each number must occur $k$ times in the sequence.

\begin{figure}[t]
\begin{lstlisting}
language Essence 1.3

given k : int(2..)
given n : int(1..)

letting seqLength be k * n

$ The sequence of numbers
find seq : sequence (size seqLength) of int(1..n)

$ The apartness constraint
such that
 $ for each number
 forAll number : int(1..n) .
  $ there exists a starting position
  $ (i.e. the first position where number occurs)
  exists start : int(1..seqLength) .
   $ the following positions all contain the same value
   $ start, start+(number+1), start+2*(number+1), ...
   forAll i : int(1..k) .
    seq(start + (i-1) * (number+1)) = number

$ symmetry breaking
such that seq(1) < seq(seqLength)

$ Each number from 1 to n appear exactly k times in seq.
$ This is an implied constraint.
such that
 forAll i : int(1..n) . |preImage(seq, i)| = k
\end{lstlisting}
\caption{\label{direct-model}\essence{} Specification: \emph{Direct} Approach to Modelling Langford's Problem.}
\end{figure}

\subsection{A \emph{Positional} Approach \label{section-positional-model}}

A second specification of Langford's Problem is what we call the \emph{Positional} approach, presented in \Cref{positional-model}. This contains a single decision variable with a domain of type function. The function maps 2-tuples, where the first component of the tuple is a number between $1$ and $n$ and the second component is the repetition index. Each 2-tuple is mapped to a position in the sequence and since these positions need to be distinct, the function is marked to be injective. \conjure{} produces an all-different constraint when refining this injective function. The apartness constraint (on lines 14--21) is written as a doubly-nested quantified expression, and for each number repetition pair it posts a binary constraint. The second constraint (on line 27) is for breaking the reflection symmetry.

\begin{figure}[t]
\begin{lstlisting}
language Essence 1.3

given k : int(2..)
given n : int(1..)

letting number be domain int(1..n)
letting repetition be domain int(1..k)
letting position be domain int(1..k*n)

$ The positions of all repetitions of all numbers
find pos : function (total, injective)
                    tuple (number, repetition) --> position

$ Occurrences of number i must be i+1 places apart.
$ So if the number 4 appears at position 3,
$ the next occurrence of it must be at position 8,
$ leaving a gap of 4 positions.
such that
 forAll i : number .
  forAll j : int(2..k) .
   pos(tuple (i,j)) = pos(tuple (i,j-1)) + i + 1

$ symmetry breaking
$ The first occurrence of the number 1 is closer to the beginning than its last occurrence is to the end.
such that
 pos(tuple (1,1)) - 1 < k*n - pos(tuple (1,k))
\end{lstlisting}
\caption{\label{positional-model}\essence{} Specification: \emph{Positional} Approach to Modelling Langford's Problem.}
\end{figure}

\subsection{Channelling the Direct and Positional Approaches}

We combine the direct and positional into a single channelled \cite{cheng1999increasing} approach. In order to do so we concatenate the two specifications into one file, excluding the common definitions like the $k$ and the $n$. We keep all the constraints regarding the problem definition in both individual specifications. The two symmetry breaking constraints are not compatible with each other, so if we post both of them we would lose solutions. We keep one or the other in our empirical evaluation.

The channelling constraints (\Cref{channelling}) are very tight by design. Specifically, they only allow one assignment to the \texttt{seq} variable for each assignment to the \texttt{pos} variable and vice versa. We validate this by removing the problem constraint from either viewpoint and enumerating all solutions.

The first channelling constraint (on lines 1--5) set values of \texttt{seq} from values of \texttt{pos}. The second channelling constraint (on lines 8--11) does the inverse. The third constraint (on lines 13--17) is not required when both problem constraints are present, but when the Positional constraints are removed they provide tighter channelling and break symmetry.

\begin{figure}[t]
\begin{lstlisting}
$ from pos to seq
such that
 forAll i : number .
  forAll j : repetition .
   seq(pos(tuple (i,j))) = i

$ from seq to pos
such that
 forAll i : int(1..k*n) .
  exists j : repetition .
   pos(tuple (seq(i),j)) = i

$ entries in pos are ordered
such that
 forAll i : number .
  forAll j : int(2..k) .
   pos(tuple (i,j-1)) < pos(tuple (i,j))
\end{lstlisting}
\caption{Channelling constraints for integrating the \emph{Direct} and \emph{Positional} models in \essence{}\label{channelling}}
\end{figure}

\section{Empirical Evaluation}

We compare the performance of solving 80 instances using the two base approaches, variations of channelled approach and a number of search strategies. The instances are generated for values of $2..6$ for $k$, and for values of $2..17$ for $n$. We use a timeout of four hours for each model-instance pair. Three instances ($(2,15), (2,16), (2,17)$) cannot be solved with any of our models, and these are excluded from our analysis. A further thirty instances are trivial (solved by one of the base models in under 5 search nodes), and these are kept in the plots. We analyse the 47 non-trivial instances in more detail. The \essence{} specifications, the parameter files, scripts for rerunning the experiments, and the raw data containing our results can be found in a source code repository hosted at \url{http://github.com/stacs-cp/ModRef2018-Langfords}.

\subsection{Comparing the Two Base Approaches}

\begin{figure}[t!]
    \centering
    \includegraphics[width=\textwidth]{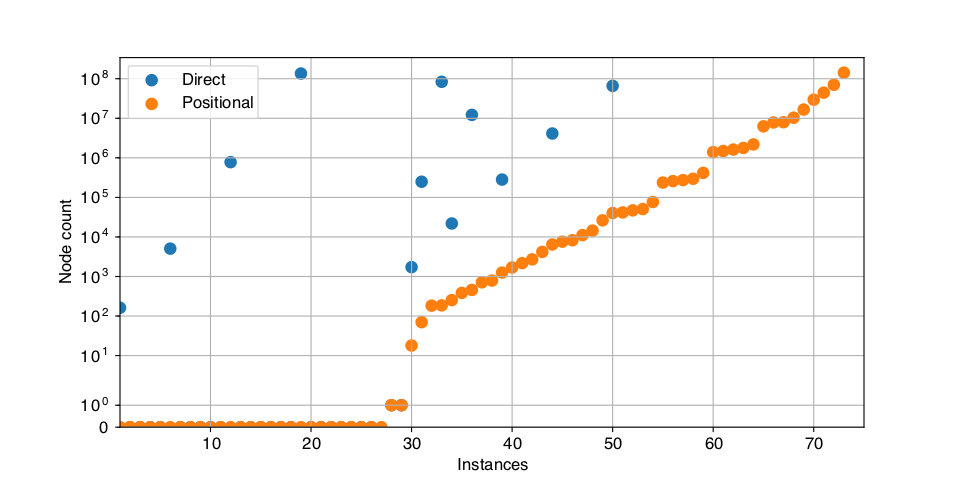}
    \caption{Search nodes with static variable ordering with the Direct model vs the Positional model. Sorted by Positional. The timed-out instances are not plotted.\label{fig:basemodels-static}}
\end{figure}

We compare the Direct and Positional approaches, using our modelling pipeline (\conjure{} 2.2.0, \savilerow{} 1.7.0) to produce constraint models suitable for input to \minion{} 1.8. We employ the default static branching order of \minion{}: the default variable ordering is by order of appearance, and the default value ordering is lexicographic. In general the positional model performs better (\Cref{fig:basemodels-static}): it is able to enumerate all solutions to all but three of out 77 instances. The Direct model times out for 40 instances and performs much worse on the small number of instances it can solve.

\begin{figure}[t!]
    \centering
    \includegraphics[width=\textwidth]{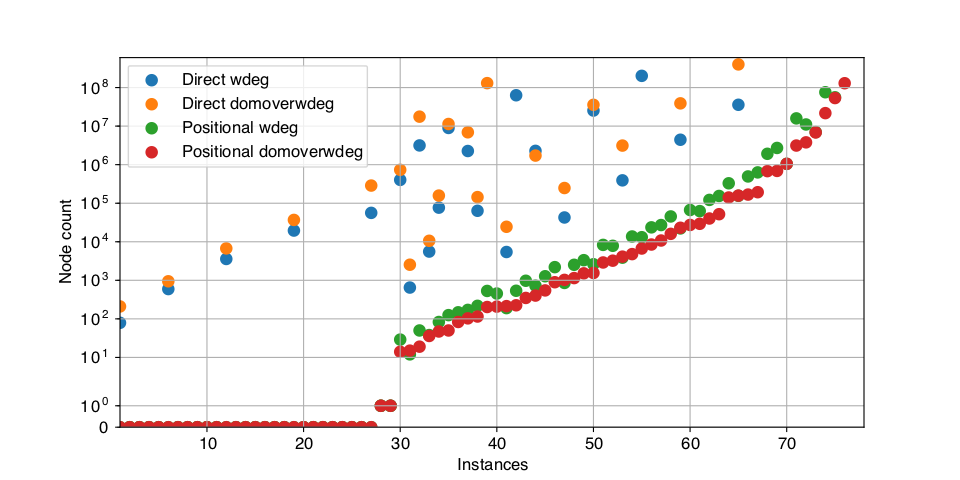}
    \caption{Search nodes with weighted degree search heuristic with the Direct model vs the Positional model. Sorted by Positional-domoverwdeg. The timed-out instances are not plotted.\label{fig:basemodels-wdeg}}
\end{figure}

A more advanced variable ordering heuristic is weighted-degree \cite{boussemart2004boosting}. \minion{} implements two variations of weighted-degree heuristics, the plain \texttt{wdeg} and another one that also takes the domain size into account: \texttt{domoverwdeg}. In \Cref{fig:basemodels-wdeg}, we see that the Positional model performs even better with one of the weighted degree heuristics (in comparison to static ordering) and it is significantly better than the Direct model.

\subsection{Channelled Approaches}

At this point the Direct approach may seem hopeless. The Positional approach gives much better performance with the static variable ordering and with weighted degree heuristics in comparison.

\begin{figure}
    \centering
    \includegraphics[width=\textwidth]{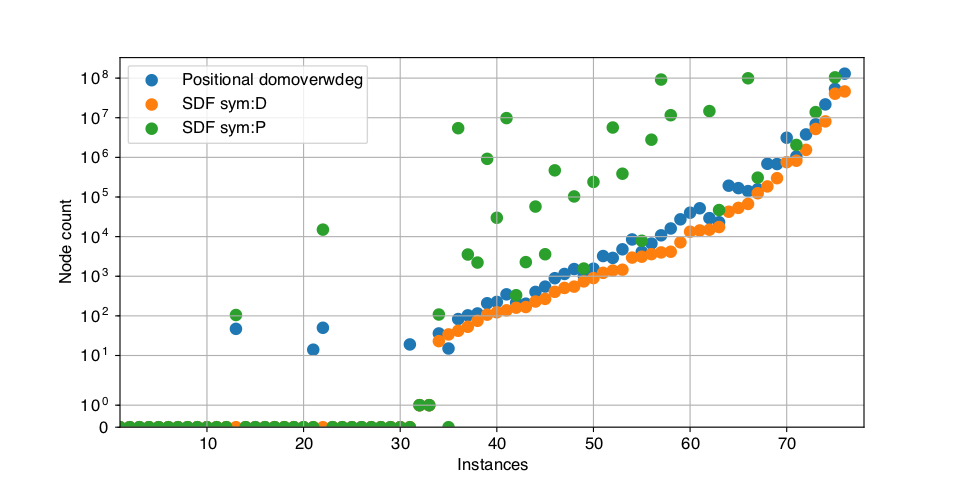}
    \caption{Search nodes with the smallest domain first heuristic on channelled models vs Positional-domoverwdeg. Sorted by Positional-domoverwdeg. The timed-out instances are not plotted.\label{fig:basemodels-ch1}}
\end{figure}

In \cite{smith2000modelling}, after considering several options the best model is found to be a channelled model, with a smallest domain first variable ordering heuristic. In that paper the authors do not break the reflection symmetry in any of their models, in this paper we would like to break this symmetry. Since each model comes with a symmetry breaking constraint and since these are incompatible, we experiment with each one separately. In \Cref{fig:basemodels-ch1}, these models are denoted by `SDF sym:P' when the symmetry breaking constraint of the Positional model is used, and by `SDF sym:D' when the Direct one is used. We compare the performance of the channelled models with the performance of the best model so far: Positional-domoverwdeg. The results are given in \Cref{fig:basemodels-ch1}.

Interestingly, the performance of the smallest domain first heuristic on the channelled model seems to depend heavily on which symmetry breaking constraint is used. The `SDF sym:P' version gives better results than Positional-domoverwdeg, whereas `SDF sym:D' is much worse overall.

\begin{figure}
    \centering
    \includegraphics[width=\textwidth]{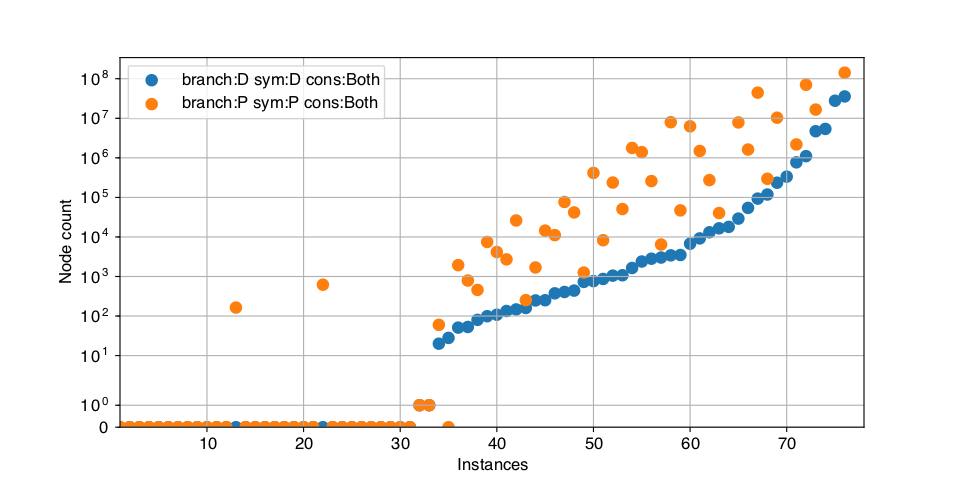}
    \caption{Comparing the number of search nodes between static variable ordering on the Direct variables vs the Positional variables. Sorted by the `branch:D sym:D' variation.\label{fig:basemodels-branchDvsP}}
\end{figure}

\Cref{fig:basemodels-branchDvsP} gives the difference between the `branch:D sym:D cons:Both' and `branch:P sym:P cons:Both' variations, where the first one uses a static variable ordering on the Direct model and the second one on the Positional model. Each model uses the corresponding symmetry breaking constraint and both sets of problem constraints. Hence, in comparison to the results given in \Cref{fig:basemodels-static}, these two variations use the same search order. However the results are inverted: branching on the Direct variables in the channelled model gives significantly better results than branching on the Positional variables. The Direct model by itself performs worse than the Positional model, yet branching on the Direct model performs significantly better than branching on the Positional model when the two models are channelled.

For reference, \Cref{table-nodes} provides detailed search node results across the best six models. It is evident that the static variable ordering on the direct variables results in the smallest number of search nodes. Further reductions are obtained by using the symmetry breaking constraints of the direct model and keeping both sets of constraints.

% \begin{landscape}
\begin{table}
    \resizebox{\textwidth}{!}{
    \begin{tabular}{C{17mm}|R{17mm}|R{18mm}|R{18mm}|R{17mm}|R{17mm}|R{17mm}}
        Instance
    &    \centering Positional dom-wdeg
    &    \centering branch:D sym:D cons:Both
    &    \centering branch:D sym:D cons:P
    &    \centering branch:D sym:P cons:Both
    &    \centering branch:D sym:P cons:P
    &    \centering SDF sym:D \tabularnewline \hline
    02\_06    &    15    &    28    &    28    &    {\bf 0}    &    12    &    34 \tabularnewline
    02\_07    &    214    &    {\bf 160}    &   {\bf  160}    &    163    &    163    &    161 \tabularnewline
    02\_08    &    1,026    &    {\bf 730}    &    {\bf 730}    &    740    &    740    &    741 \tabularnewline
    02\_09    &    4,112    &    {\bf 3,015}    &    {\bf 3,015}    &    3,075    &    3,075    &    3,111 \tabularnewline
    02\_10    &    23,064    &    {\bf 16,526}    &    {\bf 16,526}    &    16,758    &    16,758    &    17,455 \tabularnewline
    02\_11    &    157,243    &    {\bf 118,165}    &    {\bf 118,165}    &    119,791    &    119,791    &    124,268 \tabularnewline
    02\_12    &    1,050,620    &    {\bf 768,853}    &    {\bf 768,853}    &    779,204    &    779,204    &    822,842 \tabularnewline
    02\_13    &    6,843,313    &    {\bf 4,705,524}    &    {\bf 4,705,524}    &    4,753,110    &    4,753,110    &    5,202,212 \tabularnewline
    02\_14    &    53,422,836    &    {\bf 35,680,214}    &    {\bf 35,680,214}    &    36,123,258    &    36,123,258    &    40,377,181 \tabularnewline
    03\_07    &    36    &    {\bf 20}    &    24    &    24    &    27    &    23 \tabularnewline
    03\_08    &    115    &    80    &    81    &    87    &    87    &    {\bf 75} \tabularnewline
    03\_09    &    402    &    248    &    248    &    279    &    279    &    {\bf 230} \tabularnewline
    03\_10    &    1,550    &    {\bf 867}    &    {\bf 867}    &    999    &    999    &    898 \tabularnewline
    03\_11    &    6,678    &    {\bf 3,495}    &    {\bf 3,495}    &    3,766    &    3,766    &    3,609 \tabularnewline
    03\_12    &    29,239    &    {\bf 13,127}    &    {\bf 13,127}    &    14,978    &    14,978    &    14,937 \tabularnewline
    03\_13    &    140,685    &    {\bf 54,281}    &    {\bf 54,281}    &    57,563    &    57,563    &    66,819 \tabularnewline
    03\_14    &    679,085    &    {\bf 235,009}    &    {\bf 235,009}    &    257,019    &    257,019    &    298,246 \tabularnewline
    03\_15    &    3,789,394    &    {\bf 1,102,150}    &    {\bf 1,102,150}    &    1,171,173    &    1,171,173    &    1,546,663 \tabularnewline
    03\_16    &    21,707,561    &    {\bf 5,384,480}    &    {\bf 5,384,480}    &    5,758,755    &    5,758,755    &    8,079,517 \tabularnewline
    03\_17    &    129,546,336    &    {\bf 27,773,357}    &    {\bf 27,773,357}    &    28,875,464    &    28,875,464    &    46,099,072 \tabularnewline
    04\_08    &    47    &    {\bf 0}    &    20    &    40    &    45    &    {\bf 0} \tabularnewline
    04\_09    &    103    &    {\bf 53}    &    {\bf 53}    &    85    &    85    &    {\bf 53} \tabularnewline
    04\_10    &    225    &    134    &    134    &    198    &    198    &    {\bf 123} \tabularnewline
    04\_11    &    892    &    {\bf 375}    &    {\bf 375}    &    563    &    563    &    404 \tabularnewline
    04\_12    &    2,902    &    {\bf 1,075}    &    {\bf 1,075}    &    1,515    &    1,515    &    1,390 \tabularnewline
    04\_13    &    10,774    &    {\bf 2,816}    &    {\bf 2,816}    &    3,873    &    3,873    &    3,990 \tabularnewline
    04\_14    &    40,113    &    {\bf 9,186}    &    {\bf 9,186}    &    12,480    &    12,480    &    13,291 \tabularnewline
    04\_15    &    167,448    &    {\bf 29,091}    &    {\bf 29,091}    &    38,974    &    38,974    &    53,075 \tabularnewline
    04\_16    &    688,349    &    {\bf 93,089}    &    {\bf 93,089}    &    124,407    &    124,407    &    184,673 \tabularnewline
    04\_17    &    3,131,087    &    {\bf 334,480}    &    {\bf 334,480}    &    444,709    &    444,709    &    755,651 \tabularnewline
    05\_09    &    14    &    {\bf 0}    &    {\bf 0}    &    {\bf 0}    &    {\bf 0}    &    {\bf 0} \tabularnewline
    05\_10    &    50    &    {\bf 0}    &    {\bf 0}    &    61    &    64    &    {\bf 0} \tabularnewline
    05\_11    &    203    &    {\bf 108}    &    112    &    179    &    179    &    167 \tabularnewline
    05\_12    &    547    &    {\bf 251}    &    {\bf 251}    &    365    &    365    &    268 \tabularnewline
    05\_13    &    1,517    &    {\bf 440}    &    {\bf 440}    &    679    &    679    &    547 \tabularnewline
    05\_14    &    4,782    &    {\bf 1,051}    &    {\bf 1,051}    &    1,580    &    1,580    &    1,459 \tabularnewline
    05\_15    &    16,045    &    {\bf 2,391}    &    {\bf 2,391}    &    3,723    &    3,723    &    4,151 \tabularnewline
    05\_16    &    51,894    &    {\bf 6,743}    &    {\bf 6,743}    &    9,527    &    9,527    &    14,315 \tabularnewline
    05\_17    &    192,892    &    {\bf 17,974}    &    {\bf 17,974}    &    24,685    &    24,685    &    42,428 \tabularnewline
    06\_10    &    19    &    {\bf 0}    &   {\bf 0}    &    {\bf 0}    &   {\bf 0}    &    {\bf 0} \tabularnewline
    06\_11    &    83    &    51    &    52    &    85    &    89    &    {\bf 42} \tabularnewline
    06\_12    &    208    &    {\bf 99}    &    101    &    151    &    151    &    107 \tabularnewline
    06\_13    &    349    &    147    &    150    &    233    &    233    &    {\bf 138} \tabularnewline
    06\_14    &    1,142    &    {\bf 406}    &    {\bf 406}    &    605    &    605    &    507 \tabularnewline
    06\_15    &    3,232    &    {\bf 765}    &    766    &    1,203    &    1,203    &    1,218 \tabularnewline
    06\_16    &    8,475    &    {\bf 1,642}    &    {\bf 1,642}    &    2,443    &    2,443    &    2,954 \tabularnewline
    06\_17    &    27,293    &    {\bf 3,424}    &    {\bf 3,424}    &    5,165    &    5,165    &    7,170 \tabularnewline \hline
    Mean   &    4,718,175    &    {\bf 1,624,811}    &    1,624,812    &    1,672,633    &    1,672,633    &    2,207,366 \tabularnewline
    Sum    &    221,754,209    &    {\bf 76,366,120}    &    76,366,156    &    78,613,734    &    78,613,761    &    103,746,215 \tabularnewline
    \end{tabular}
    }
\vspace{5mm}
\caption{Search nodes for our best six models of Langford's Problem. \label{table-nodes}}
\end{table}
% \end{landscape}

\subsection{Comparison against \texttt{BC\_MINISAT\_ALL}}

We compare the performance of one of the best CP models to that of an AllSAT solver, BC\_MINISAT\_ALL \cite{toda2016implementing}. We use \savilerow{} to produce a SAT encoding for the Direct model, the Positional model, and a channelled model with the symmetry breaking constraint from the Positional model. We only experiment with these three variations for SAT, since variable ordering heuristics cannot be specified like they can in a constraint solver.

The SAT encoding for the Direct model times out on eleven of 77 instances with a 4-hour timeout, and the Positional model times out on four. The channelled model performs better than either model in SAT as well, it solves all instances in 13,266 seconds in total. In comparison the channelled constraint model with the `branch:D sym:P cons:P' variation (branching on Direct variables, symmetry breaking using the Positional constraint, and only using the Positional problem constraints) solves all instances in 7,408 seconds in total. 

\section{Conclusion}

In this paper we have studied the performance of two different viewpoints of Langford's Problem, both separately and channelled together. Earlier work has demonstrated the utility of the channelled approach, but in contrast to previous findings, we found that the most effective search strategy to exploit the channelled models is a static variable ordering on the viewpoint that, in isolation, is weaker. Our conjecture is that, through the channelling constraints, this static variable ordering produces a high quality dynamic variable ordering on the second viewpoint. 

In future work, we will investigate whether this finding can be translated to other problem classes and whether our findings also hold with other constraint solvers.

\subsubsection*{Acknowledgements}

This work was supported by EPSRC EP/P015638/1. We thank our anonymous reviewers for helpful comments.

\bibliographystyle{splncs04}
\bibliography{langford}

\begin{thebibliography}{10}
\providecommand{\url}[1]{\texttt{#1}}
\providecommand{\urlprefix}{URL }
\providecommand{\doi}[1]{https://doi.org/#1}

\bibitem{akgun2014extensible}
Akg{\"u}n, {\"O}.: Extensible automated constraint modelling via refinement of
  abstract problem specifications. Ph.D. thesis, University of St Andrews
  (2014)

\bibitem{akgun2013automated}
Akg{\"u}n, {\"O}., Frisch, A.M., Gent, I.P., Hussain, B.S., Jefferson, C.,
  Kotthoff, L., Miguel, I., Nightingale, P.: Automated symmetry breaking and
  model selection in {C}onjure. In: International Conference on Principles and
  Practice of Constraint Programming. pp. 107--116. Springer (2013)

\bibitem{akgun2014breaking}
Akg{\"u}n, {\"O}., Gent, I.P., Jefferson, C., Miguel, I., Nightingale, P.:
  Breaking conditional symmetry in automated constraint modelling with
  {C}onjure. In: ECAI. pp.~3--8 (2014)

\bibitem{akgun2011extensible}
Akg{\"u}n, {\"O}., Miguel, I., Jefferson, C., Frisch, A.M., Hnich, B.:
  Extensible automated constraint modelling. In: Proceedings of theTwenty-Fifth
  AAAI Conference on Artificial Intelligence. pp. 4--11. AAAI Press (2011)

\bibitem{boussemart2004boosting}
Boussemart, F., Hemery, F., Lecoutre, C., Sais, L.: Boosting systematic search
  by weighting constraints. In: ECAI. vol.~16, p.~146 (2004)

\bibitem{cheng1999increasing}
Cheng, B., Choi, K.M.F., Lee, J.H.M., Wu, J.: Increasing constraint propagation
  by redundant modeling: an experience report. Constraints  \textbf{4}(2),
  167--192 (1999)

\bibitem{frisch2005essence}
Frisch, A.M., Grum, M., Jefferson, C., Hern{\'a}ndez, B.M., Miguel, I.: The
  {E}ssence of {E}ssence. Modelling and Reformulating Constraint Satisfaction
  Problems pp. 73--88 (2005)

\bibitem{frisch2007design}
Frisch, A.M., Grum, M., Jefferson, C., Hern{\'a}ndez, B.M., Miguel, I.: The
  design of {E}ssence: A constraint language for specifying combinatorial
  problems. In: IJCAI. pp. 80--87 (2007)

\bibitem{frisch2008ssence}
Frisch, A.M., Harvey, W., Jefferson, C., Mart{\'\i}nez-Hern{\'a}ndez, B.,
  Miguel, I.: {Essence}: A constraint language for specifying combinatorial
  problems. Constraints  \textbf{13}(3),  268--306 (2008)

\bibitem{gent2006minion}
Gent, I.P., Jefferson, C., Miguel, I.: Minion: A fast scalable constraint
  solver. In: ECAI. vol.~141, pp. 98--102 (2006)

\bibitem{hnich2004dual}
Hnich, B., Smith, B.M., Walsh, T.: Dual modelling of permutation and injection
  problems. Journal of Artificial Intelligence Research  \textbf{21},  357--391
  (2004)

\bibitem{hnich2003channel}
Hnich, B., Walsh, T.: Why channel? multiple viewpoints for branching
  heuristics. Modelling and Reformulating Constraint Satisfaction Problems
  p.~94 (2003)

\bibitem{nightingale2014automatically}
Nightingale, P., Akg{\"u}n, {\"O}., Gent, I.P., Jefferson, C., Miguel, I.:
  Automatically improving constraint models in {S}avile {R}ow through
  associative-commutative common subexpression elimination. In: CP. pp.
  590--605. LNCS 8656, Springer (2014)

\bibitem{sr-journal-17}
Nightingale, P., Akg\"un, O., Gent, I.P., Jefferson, C., Miguel, I., Spracklen,
  P.: Automatically improving constraint models in {Savile Row}. Artificial
  Intelligence  \textbf{251},  35--61 (2017).
  \doi{10.1016/j.artint.2017.07.001}

\bibitem{nightingale2015automatically}
Nightingale, P., Spracklen, P., Miguel, I.: Automatically improving {SAT}
  encoding of constraint problems through common subexpression elimination in
  {S}avile {R}ow. In: CP. pp. 330--340. LNCS 9255, Springer (2015)

\bibitem{smith9comparing}
Smith, B.M.: Comparing dual viewpoints in permutation problems. Constraint
  Modelling and Reformulation (ModRef’09) p.~147

\bibitem{smith2000modelling}
Smith, B.M.: Modelling a permutation problem  (2000)

\bibitem{smith2001dual}
Smith, B.M.: Dual models of permutation problems. In: International Conference
  on Principles and Practice of Constraint Programming. pp. 615--619. Springer
  (2001)

\bibitem{toda2016implementing}
Toda, T., Soh, T.: Implementing efficient all solutions sat solvers. Journal of
  Experimental Algorithmics (JEA)  \textbf{21},  1--12 (2016)

\bibitem{walsh2001permutation}
Walsh, T.: Permutation problems and channelling constraints. In: International
  Conference on Logic for Programming Artificial Intelligence and Reasoning.
  pp. 377--391. Springer (2001)

\end{thebibliography}

\end{document}